\documentclass[letterpaper, 10 pt, conference]{ieeeconf}  

\IEEEoverridecommandlockouts                              

\usepackage{amsmath,amssymb,amsfonts}
\usepackage{graphicx}
\usepackage{setspace}
\usepackage{subcaption}
\usepackage{float}
\usepackage{multirow}
\usepackage{algorithm}
\usepackage{algpseudocode}
\usepackage{breqn}
\usepackage{array}
\usepackage{flushend}
\usepackage{url}
\usepackage{soul}
\usepackage{color}
\usepackage{marginnote}
\usepackage[colorlinks,bookmarksopen,bookmarksnumbered,citecolor=red,urlcolor=red]{hyperref}

\def\BibTeX{{\rm B\kern-.05em{\sc i\kern-.025em b}\kern-.08em
    T\kern-.1667em\lower.7ex\hbox{E}\kern-.125emX}}

\title{\LARGE \bf
Mid-flight Propeller Failure Detection and Control of Propeller-deficient Quadcopter using Reinforcement Learning
}

\author{Rohitkumar Arasanipalai$^{\dagger,1}$, Aakriti Agrawal$^{\dagger,1}$ and Debasish Ghose$^{2}$
\thanks{*Supported by EPSRC-GCRF (EP/P023839X/1) and RBCCPS, IISc.}
\thanks{$^{\dagger}$ Equal Contribution}%
\thanks{$^{1}$R. Arasanipalai and A. Agrawal are Research Assistants, and D. Ghose is a professor, at the Guidance, Control, and Decision Systems Laboratory (GCDSL), Dept. of Aerospace, Indian Institute of Science, Bangalore, India. Emails:        {\tt\small rohitkumar97@gmail.com}; {\tt\small juhi05aakritiagrawal@gmail.com}; {\tt\small dghose@iisc.ac.in}}%
}

\begin{document}
\maketitle
\thispagestyle{empty}
\pagestyle{empty}

\begin{abstract}

Quadcopters can suffer from loss of propellers in mid-flight, thus requiring a need to have a system that detects single and multiple propeller failures and an adaptive controller that stabilizes the propeller-deficient quadcopter. This paper presents reinforcement learning based controllers for quadcopters with 4, 3, and 2 (opposing) functional propellers. The paper also proposes a neural network based propeller fault detection system to detect propeller loss and switch to the appropriate controller. The simulation results demonstrate a stable quadcopter with efficient waypoint tracking for all controllers. The detection system is able to detect propeller failure in a short time and stabilize the quadcopter.

\end{abstract}

\section{INTRODUCTION}\label{section:intro}

Autonomous quadcopter UAVs often suffer from loss of one or multiple propeller(s) mid-flight \cite{Birk,rescue}. Unless the controller is robust enough to enable flight in propeller-deficient condition, the UAV crashes, causing damage to itself as well as the surroundings. This paper proposes a fault detection (FD) system to detect propeller failure mid-flight and reinforcement-learning (RL) based controllers to control the propeller-deficient quadcopter.

\begin{figure}
  \centering
\includegraphics[width=0.75\linewidth,trim={0cm 0cm 0cm 0cm},clip]{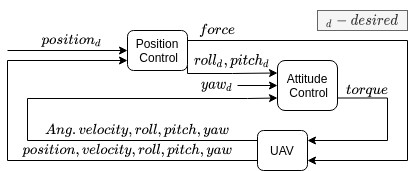}
\caption{The inner and outer control loop of a quadcopter.}
\label{fig:2_loops}
\end{figure}

Controllers used in quadcopters consists of two loops in the control model (Fig. \ref{fig:2_loops}; See Ref. \cite{Cao2016InnerOuterLC}); the outer loop for waypoint tracking and inner one for stability. The decision-making system, proposed in this paper, has a similar structure, but with a RL agent in the outer loop, and a PD controller in the inner loop. Previous work on propeller loss scenarios \cite{control_main} have developed separate control systems based controllers for 3, 2 (opposing), and 1 propeller lost quadcopters. We have also done the same for 2 (opposing) and 1 propeller loss scenarios, but by using RL, which allows learning more complex behaviour and is adaptable to different conditions. Earlier methods \cite{RL_main}, \cite{Waslander2005MultiagentQT} based on RL were developed for quadcopters with no propeller failure. 

Although \cite{control_main} designed controllers for quadcopters with propeller failure, it lacked an online FD system to switch between controllers during flight. We propose a method using deep learning that detects specific propeller loss using information collected from on-board sensors. {No additional sensors are used, thus avoiding addition of any extra weight to the quadcopter.} The two systems are combined to achieve both propeller failure detection and controller switching in mid-flight. We show that RL controllers are capable of waypoint tracking even with multiple lost propellers, thus enabling the quadcopter to complete the mission.

The RL setup consists of two major components, an agent and an environment. The agent is the decision-maker which gives control commands to the four motors and the environment is the quadcopter which the agent is acting on. The quadcopter changes its position and orientation when acted upon by the agent. Most RL algorithms follow a similar pattern. First, the environment passes the initial state to the agent, which then acts in order to proceed to the next state. The environment then returns the new state along with the reward of the previous action. Based on the reward, the agents learn which action-state pair maximize the rewards. This loop continues until the terminal state is reached.


{We use a model-free RL algorithm with deterministic policy, as we do not intend to learn the complex dynamics of the environment. This is unlike model-based RL algorithms which need to learn the complete state transition probability from the pair of current state and action to the next state.}

\begin{figure}
  \centering
  \includegraphics[width=0.65\linewidth,trim={0cm 0cm 0cm 0cm},clip]{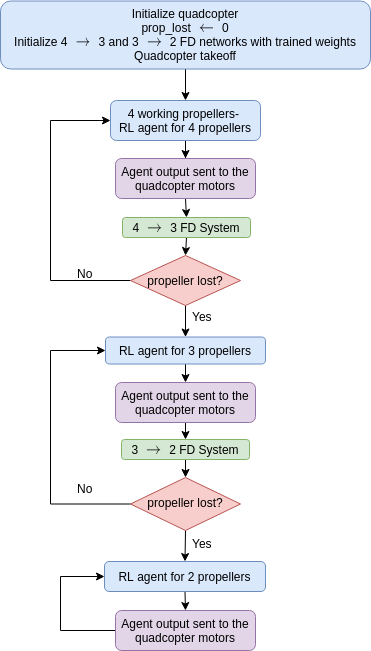}
\caption{Fault detection system and RL controllers}
\label{fig:system_flowchart}
\end{figure}

The organization of the paper is as follows: Section \ref{section:rr} discusses previous related work, Section \ref{section:method1} describes the proposed RL decision making system for quadcopter control with failed propellers. Section \ref{section:method2} describes the fault-detection system, and finally Section \ref{section:method3} combines these two systems together. A thorough comparison of results is done in Section \ref{section:result} and conclusions are drawn in Section \ref{section:conc}.

\section{Related Research}\label{section:rr}

In \cite{Waslander2005MultiagentQT}, model-based RL was used to find an optimal policy for the altitude control loop, yielding a stable controller for a quadcopter. In \cite{RL_main}, a PD controller was used for stability and an RL agent for waypoint tracking. The RL agent also learns many different complexities such as maintaining orientation and stability. In \cite{RL_attitude}, only attitude control was achieved using RL, focusing on the inner loop as a first step. These systems do not address propeller-loss conditions.

Fault-tolerant control systems for single propeller loss has been proposed in \cite{fault_tol,fault_tole}, which focus on the angular velocity along the vertical axis and then carry out path following using three propellers. This idea was used in \cite{control_main}, where they have derived the control equation and constraints required to fly a quadcopter with only 3, 2 or 1 functioning propeller. They showed take-off and waypoint tracking but not mid-flight propeller loss detection or switching between controllers. Similarly, in \cite{prop_loss_IJRR}, hover conditions are derived for 3, 2 or 1 propeller lost scenarios. In \cite{FD_FTC}, a combined fault detection and controller was developed for loss of a single actuator only. The papers (\cite{fault_tol,fault_tole, prop_loss_IJRR,FD_FTC}) use control theoretic methods and not RL. Fault detection, diagnosis, and control for unmanned rotorcraft systems, from a control theoretic perspective, are surveyed in \cite{review_FD}. Some related work monitors structural health in real-time \cite{yap}, and off-line propeller fault detection using neural networks \cite{prop_sound} and  spectral analysis \cite{fault_prop} for operational check before flight. 

In contrast, we present a learning based combined FD system for multiple propeller failures and an RL adaptive controller to stabilize the quadcopter post-failure.


\section{Control of quadcopter with 4, 3, and 2 functional propellers using RL}\label{section:method1}

Three different RL-based controllers for no propeller loss, 1 propeller loss, and 2 propeller loss, are designed. A FD system using recurrent neural networks (RNNs) to identify the propeller(s) that have failed mid-flight, is also implemented. After fault detection, we switch to the appropriate controller mid-flight, thus enabling control and waypoint tracking even after the loss of 1 or 2 propellers. A schematic of our system can be seen in Fig. \ref{fig:system_flowchart}. Out of the two control loops in the quadcopter controller, shown in Fig. \ref{fig:2_loops}, the inner loop is PD control based and the outer loop is RL-control based.

\subsection{Quadcopter control using RL}
\normalmarginpar
\begin{figure}
  \centering
  \includegraphics[width=0.5\textwidth,trim={0cm 6cm 0cm 0cm},clip]{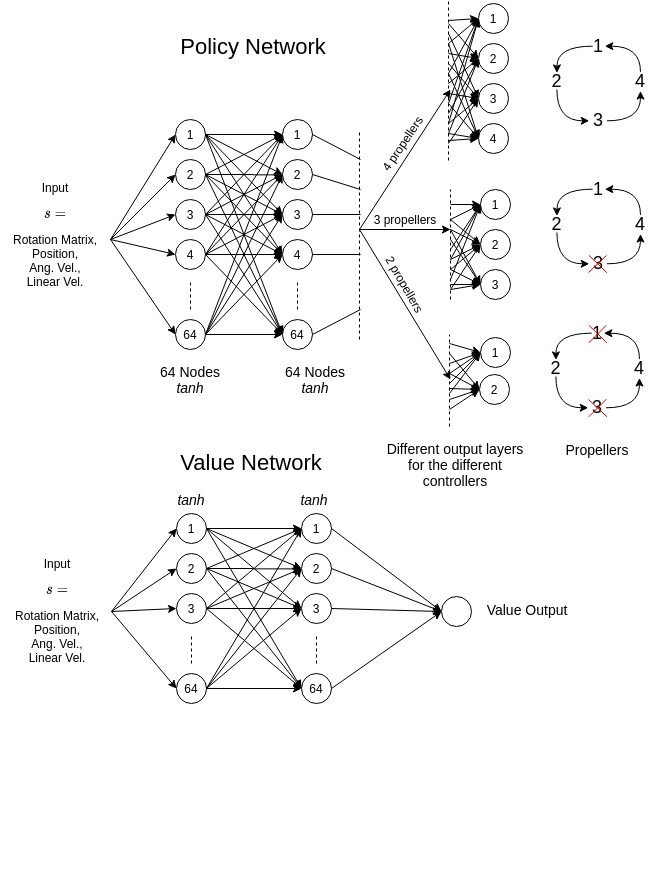}
\caption{The policy network and the value function network.  The cyclic allotment of the output is also shown.}
\label{fig:agent_network}
\end{figure}


There exist several policy optimization algorithms like PPO (Proximal Policy Optimization)  \cite{PPO}, DDPG (Deep Deterministic Policy Gradient) \cite{DDPG} and TRPO (Trust Region Policy Optimization) \cite{TRPO}. Among these, DDPG, has convergence issues and TRPO is complex and computationally intensive. Hence, we selected PPO which has been found to be simpler and computationally efficient \cite{RL_main}.

{The agents consists of two networks for training, a value network, and a policy network. Both networks have 2 hidden layers of 64 nodes with tanh activation function.}{ The network architecture is given in Fig. \ref{fig:agent_network}}. The input is the quadcopter state which is an 18-element vector $ s$ { (all quantities are defined in the inertial frame)}:
\begin{align}
    s = [R_{flat}, x, y, z, v_x, v_y, v_z, w_x, w_y, w_z]
    \label{eq:quad_state}
\end{align}
where, $R_{flat}$ is the flattened form of quadcopter's rotation matrix, ($x$, $y$, $z$) is the quadcopter position, ($v_{x}$, $v_{y}$, $v_{z}$) are the  linear, and ($w_{x}$, $w_{y}$, $w_{z}$) are the angular, velocities of the quadcopter. The output is an $n$-element vector where $n$ is the number of functional propellers. We used Huber loss function \cite{huber} for loss calculation of value network and standard gradient descent \cite{SGD} for the policy network.

a2-c3
{ In order to ensure safety during learning, the episode (trajectory) that comes close to violating a safety constraint can be terminated or given a large cost, as in \cite{Cheng}.}

The value function $V(s|\eta)$ is trained using Monte-Carlo (MC) samples that are obtained from the on-policy trajectories. Terminal value, that is, the tail cost of the trajectory, is taken from the current value function. 
\begin{equation}
    v_{t} = \sum_{i=t}^{T-1} \gamma^{i-1}r_{i} + \gamma^{T-t}V(s_{T}|\eta) \label{eq:1}
\end{equation}
where, $\eta$ are the parameters of the approximated value function, $T$ is the length of the trajectory, $\gamma$ is the discount factor and $r$ is reward as given in (\ref{eq:reward}) below.

\textbf{Policy Optimization:} { As mentioned, we have used PPO, a policy optimization based algorithm, given in Algorithm \ref{po}. The details of PPO can be found in \cite{PPO}. We define our policy as $\pi(s|\theta)$ with parameters $\theta$}.

\begin{algorithm}
\caption{Policy Optimization}\label{po}
\begin{algorithmic}[1]
    \State Initialize parameters for $V(s|\eta)$ and $\pi(s|\theta)$
    \State \textbf{while} $j$ = 1,2,3.. until convergence \textbf{do}
    \State\hspace{\algorithmicindent} Collect data according to \textbf{Exploration Strategy}
    \State\hspace{\algorithmicindent} Compute MC estimate of $v^{p}_{i}$ using (\ref{eq:1}) 
    \State\hspace{\algorithmicindent} Update $V(s|\eta)$ $n_{v}$ times using Huber loss.
    \State\hspace{\algorithmicindent} Update $\pi(s|\theta)$ once using standard gradient descent.
    \State \textbf{end while}
\end{algorithmic}
\end{algorithm}

{ The quadcopter simulation environment\footnote{The simulation software used is the \href{https://github.com/leggedrobotics/raisimLib}{RAISIM} physics simulator and the \href{https://github.com/leggedrobotics/raisimGym}{RaisimGym} quadcopter environment (\href{https://github.com/leggedrobotics/raisimGym}{https://github.com/leggedrobotics/raisimGym}). The interface to this environment is similar to the OpenAI gym environment and has a Python interface.} takes actions as input and returns the updated state of the quadcopter along with the rewards. These states and rewards are stored and used to train the RL network. The actions are given by the partially trained RL network. Multiple environments can be run in parallel which helps in exploration when running the Monte-Carlo simulations. Propeller failure was simulated as mentioned in Section \ref{section:IV_C}.}

During policy optimization, the policy is trained with the origin of the inertial frame as the target waypoint. During operation, the origin of the inertial frame is shifted to the target waypoint. This is done so that the policy need not be explicitly trained on waypoint tracking. The quadcopter is initialized in a random normally distributed state  (that is, random position, orientation, angular velocity, and linear velocity) with a reasonable bound such that we can easily explore the feasible state space. {  We have initialized the various parameters by sampling from a truncated Gaussian distribution with limits of $[-3\sigma, +3\sigma]$ as follows: Position $\sim N(0,1)$ (in meters); Orientation is a quaternion vector with 4 elements, each of which are sampled from $N(0,1)$ and normalized; Angular velocity $\sim N(0,5)$ (in rad/sec); Linear velocity $\sim N(0,5)$ (in meters/sec). We use large limits to train the agent on extreme conditions and thus ensure higher levels of robustness.} Each epoch of the training is done on 500 trajectories, each of which has 500 timesteps. {The control frequency is 100 Hz and therefore, each trajectory is of 5 seconds}. { The 3 controllers (4, 3 and 2 propeller controllers) were trained for 4500 epochs. The stopping criteria was value loss $<$ 0.0001. Each training session was for 14 hours on an Nvidia Geforce 960M with 2GB memory.}

As mentioned, the quadcopter uses a PD controller to maintain stability. { The motor output of the RL network is converted into force and torque using quadcopter dynamics which are then added to the force and torque output of the PD controller, respectively. The simulator then applies the total force and torque on the quadcopter model.} The PD controller alone is insufficient, but helps in avoiding extreme movements and therefore, aids in stabilizing the learning process. Without it, the quadcopter simply goes out of bounds due to the random state initialization. The PD controller is given as \cite{RL_main} $\tau_{b} = k_{p}R^{T}q + k_{d}R^{T}w$ where, $\tau_{b}$ is the virtual torque produced on the main body as a result of the thrust forces, { $q$ is the euler orientation vector, $R$ is the rotation matrix} and $w$ is the angular velocity. { The values of $k_{p}$ and $k_{d}$ are $-0.2$ and $-0.06$ for the $x$ and $y$ direction, and $-0.033$ and $-0.01$ for the $z$ direction, which is one-sixth of the values used for $x$ and $y$ directions, as prescribed in \cite{RL_main}.} Reward at any time $t$ is defined as, 
\begin{multline}
    r_{t} = 2\times10^{-3}||p_{t}|| + 1\times10^{-4}||w_{t}|| + 5\times10^{-4}||\alpha_{t}||
    \label{eq:reward}
\end{multline}
where, $p_{t}$ and $w_{t}$ are the current position and angular velocities, respectively. The angle between the quadcopter's vertical axis and $z$-axis of the inertial frame is $\alpha_t$. Position has a high coefficient since waypoint tracking is the priority of the RL agent. Discount factor $\gamma = 0.99$.

We have used the same network to train different controllers for no propeller loss, 1 propeller loss, 2 propeller loss with some important modifications for each case. These are discussed in subsequent sections.

\subsection{Quadcopter control with no propeller loss}
The RL agent is similar to that defined above. The output of the agent is 4 action values which are motor speeds ($w$ = [$w_{1}$, $w_{2}$, $w_{3}$, $w_{4}$]). These values are combined with the values from the PD controller to give the final output. 

\subsection{Quadcopter control with one propeller loss}
\begin{figure}
  \centering
  \includegraphics[width=\linewidth]{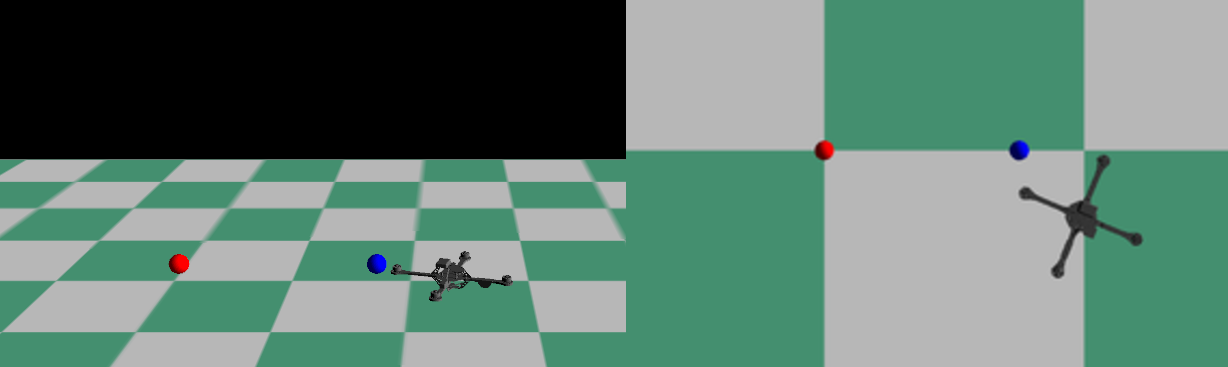}
  \caption{ Simulation of the quadcopter with 2 failed propellers}
  \label{fig:both_dots}
\end{figure}

For the quadcopter to be controllable, it was shown in \cite{control_main} that $n_{z} \neq 0$, where $n =$ ($n_{x}, n_{y}, n_{z}$) is the unit vector governed by the differential equation: $\dot{n} = -w^{B} \times n$ (where $w^B$ is the quadcopter's angular velocity in the body reference frame). Hence, we conclude that the quadcopter should rotate about a body axis whose vertical component is not 0. Fig. \ref{fig:both_dots} shows the simulation of the quadcopter with 2 propellers lost. The red dot shows the first target waypoint and the blue dot shows the shifted waypoint.

The RL agent gives 3 action outputs ($w = [w_{i}, w_{i+1}, w_{i+2}]$). { These outputs are assigned to the propeller starting from the first working propeller in a cyclic manner. Fig. \ref{fig:agent_network} shows this for 1 and 2 propeller loss. Here, output 1 of the 3 propeller network would be assigned to propeller 1, output 2 would be assigned to propeller 2 and output 3 would be assigned to propeller 4.}

\subsection{Quadcopter control with two opposite propeller loss}
Similar to the previous section, following \cite{control_main}, $n_{z} = 1$ to control the quadcopter with two opposite propellers lost and $l \neq 0$. Note that we are not considering the failure of two adjacent propellers, which is an unsolved problem in the literature. The RL agent gives two outputs ($w = [w_{i}, w_{i+1}]$), which acts on the 2 opposing functional motors. {  These outputs are assigned in a similar cyclic manner as mentioned in the previous subsection and shown in Fig. \ref{fig:agent_network}.}

\section{Fault-detection using neural network}\label{section:method2}
Recurrent Neural Networks (RNNs) \cite{rnn} have been used to exploit the temporal relationship between elements of a sequence by recursively ('unrolling') processing each element. For 1 and 2 lost propellers, the quadcopter exhibits unique set of states at each timestep, which can be classified by an RNN. One problem with RNNs is the exponential growth or decay in the gradient vector for long sequences during training, which prohibits learning long-distance correlations in the sequence. Therefore, we have used LSTM (Long Short Term Memory) \cite{lstm} which does not have the above issue.

\subsection{Fault-detection}
For the remainder of this paper, the nomenclature of $m\rightarrow n$ is used to denote failure cases, where $m$ is the number of functional propellers before fault occurs, and $n$, the number of functional propellers after the fault. For example, $4\rightarrow3$ denotes the quadcopter going from 4 functional propellers to 3 functional propellers after a propeller failure.

The task is to map the quadcopter states (as defined in (\ref{eq:quad_state})) to possible propeller failure outcomes. There are 5 possible outcomes when going from 4 functional propellers to 3 functional propellers: no propeller lost or one of the 4 propellers lost. Then, the elements of the output vector $Q_{4\rightarrow3} \in \mathbb{R}^{5}$ denote probability of of one of the five outcomes. 

When going from 3 functional propellers to 2 functional propellers, we have considered only the opposite propeller failure option. Therefore, $Q_{3\rightarrow2} \in \mathbb{R}^{2}$. We can now denote the neural network as a function $f$ trained to map:
\begin{align}
    f:[s_{t-T},s_{t-T+1},...,s_{t}] \mapsto Q_t
    \label{eq:network_equation}
\end{align}

\subsection{Network architecture}

\begin{figure}
  \centering
  \includegraphics[width=\linewidth,trim={0cm 0cm 0 0},clip]{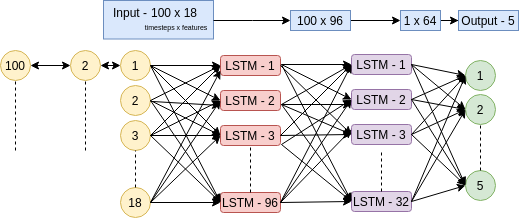}
\caption{FD network for $4\rightarrow3$ detection system. }
\label{fig:4_3_network}
\end{figure}

\subsubsection{4 $\rightarrow$ 3 fault-detection (FD) network}
It consists of 96 LSTM cells in the first layer and 64 in the second layer (See Fig. \ref{fig:4_3_network}). The second layer passes into the output feedforward layer which has 5 nodes for the 5 possible classes/outcomes. The optimizer used is Stochastic Gradient Descent (SGD) \cite{SGD} with momentum. Momentum allows the network to converge over a wide range of learning rates \cite{mom} thus allowing us to prototype multiple networks quickly.
\begin{figure}
  \centering
  \includegraphics[width=\linewidth,trim={0cm 0cm 0 0},clip]{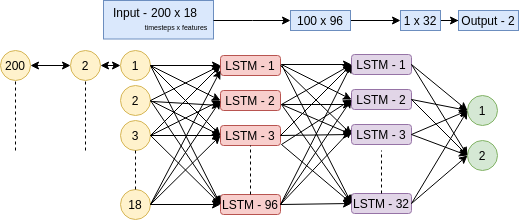}
\caption{FD network for $3\rightarrow2$ detection system. }
\label{fig:3_2_network}
\end{figure}
\subsubsection{3 $\rightarrow$ 2 fault-detection (FD) network}
It consists of 96 LSTM cells in the first layer and 32 in the second layer (See Fig. \ref{fig:3_2_network}). It passes through the feedforward layer with 2 nodes for the 2 possible classes. The optimizer used is Adam \cite{adam}.

The training parameters are listed in Table \ref{tab:optimizer_specs}. { Note that we trained both the networks using both SGD and Adam.The final optimizer chosen for each network was the one that gave better final training accuracy.} We chose a computationally cheap network with only two LSTM layers for two reasons. Firstly, the network would ultimately need to run in real-time on the drone, which generally have light CPU/GPU. Secondly, even small networks can learn very complex contextual information. Our network gave highly accurate results and thus there was no need for a complicated network.

\begin{table}
  \centering
  \resizebox{\columnwidth}{!}{%
    \begin{tabular}{|l|l|l|}
        \hline
        & \text{$4\rightarrow3$ Transition Network} & \text{$3\rightarrow2$ Transition Network}\\
        \hline
        \text{Learning Rate} & $10^{-4}$ & $10^{-4}$\\
        \text{Momentum} & 0.9 &  - \\
        \text{No. of Training Epochs} & 95 & 416 \\
        \text{Final Training Accuracy} & 97\% & 92\% \\
        \hline
    \end{tabular}
    }
    \caption{Training parameters for the 2 networks}
        \label{tab:optimizer_specs}
\end{table}

We have used a window size ($T$) of 100 for $4\rightarrow3$ FD network and 200 for $3\rightarrow2$ FD network as shown in Fig. \ref{fig:4_3_network} and \ref{fig:3_2_network}, respectively. Using a lower window size of 50 and 100 for the two networks, respectively, gave lower accuracy in the predictions because we needed more timesteps to establish the relationship between quadcopter behavior and propeller loss. Larger window length is likely to increase the fault-detection time as well which is undesirable.

The first 150 timesteps for the $4\rightarrow3$ FD network and 250 timesteps for the $3\rightarrow2$ FD network are skipped because the quadcopter is initialized in a random state and we do not want the network to learn this initial erratic behaviour that occurs until stabilization. Instead, we want the network to learn the erratic behaviour when a propeller loss occurs. Thus, the RNN does not need to learn to differentiate between the two behaviors, as the initial erratic behavior is unlikely to be encountered in actual scenarios.

Let the $4\rightarrow3$ FD network, with a window of 100 and starting from 151-st sample, be represented as function $f_{4\rightarrow3}$. Following (\ref{eq:network_equation}), the input and output can be related as,
\begin{align}
    f_{4\rightarrow3}(\text{$s_{t-100}$, $s_{t-99}$, ..., $s_{t}$}) = Q_{4\rightarrow3, t}, \text{where $t>150$}
\end{align}
Similarly, for the $3\rightarrow2$ FD network, with a window of 200 and starting from 251-st sample are related as,
\begin{align}
    f_{3\rightarrow2}(\text{$s_{t-200}$, $s_{t-199}$, ..,$s_{t}$}) = Q_{3->2, t}, \text{where $t>250$}
\end{align}

{\subsection{Data collection}
\label{section:IV_C}
 
For the $4\rightarrow3$ FD network, the 4 propeller controller was used to control the quadcopter even with a propeller loss mid-flight. The data was collected from 500 simulations of the propeller loss scenario and the network was trained on this data. For the $3\rightarrow2$ FD network, the same procedure was followed but the quadcopter started with 3 working propellers and lost 1 propeller while it was in-flight and being controlled by the 3 propeller controller. In simulation propeller failure was implemented by turning off either one or two of the propellers. The data collected were of position, orientation, angular velocity and linear velocity. The labels were collected from the number of working propellers and encoded using one-hot encoding.}

\section{Complete System}
\label{section:method3}
The complete system, combining the fault detection and RL controller, was shown in Fig.  \ref{fig:system_flowchart}. We assume that the quadcopter starts with 4 working propellers. The RL agent based controller for 4 propellers is engaged. The $4\rightarrow3$ FD network continuously checks for propeller failure at every loop. There are four cases here for each of the four propellers of the quadcopter, that is, either propeller 1 fails or propeller 2 fails and so on. Once a propeller failure is detected, the same is updated and the controller for 3 propellers is engaged. From this point onward, the $3\rightarrow2$ FD network takes over and checks for the second propeller loss. Similar to the above, if it encounters the second (opposing)
propeller failure, it switches to 2 propeller controller. 


\subsection{Removing offset}\label{subsec:off}
Deep-RL suffers from the bias vs. variance paradigm.{ PPO algorithm has less variance but suffers from large bias.  We also observe a small, but constant offset between the quadcopter's position and the required position. One likely reason could be the bias in the function-approximator.} Since it is a constant offset we handle it using a moving average filter with a window of 15 time-steps, and average the quadcopter's actual position within this window. {This could also have been done by computing the integral error.} Since our required position is the origin of the inertial frame ($[0,0,0]$), and our model is trained for that position, we add the moving average value to the quadcopter's actual position. 

Let the quadcopter's actual position at the $t-th$ time instant be $x(t), y(t), z(t)$, and let  $\overline{x}(t)$, $\overline{y}(t)$, $\overline{z}(t)$ be the moving average in $x$, $y$, and $z$ direction. Then, 
                
\begin{align}
\overline{p}(t) &= \frac{\sum_{i=t}^{t+15} p_{i-15}}{15} \text{   where},\text{ }p = (x,~y,~z).
\end{align}
Now add $\overline{p}$ offset to the quadcopter's actual position, to make it the quadcopter's observed position.

\begin{figure}
\begin{minipage}{0.5\textwidth}
\begin{subfigure}{\textwidth}
  \centering
  \includegraphics[width=\linewidth]{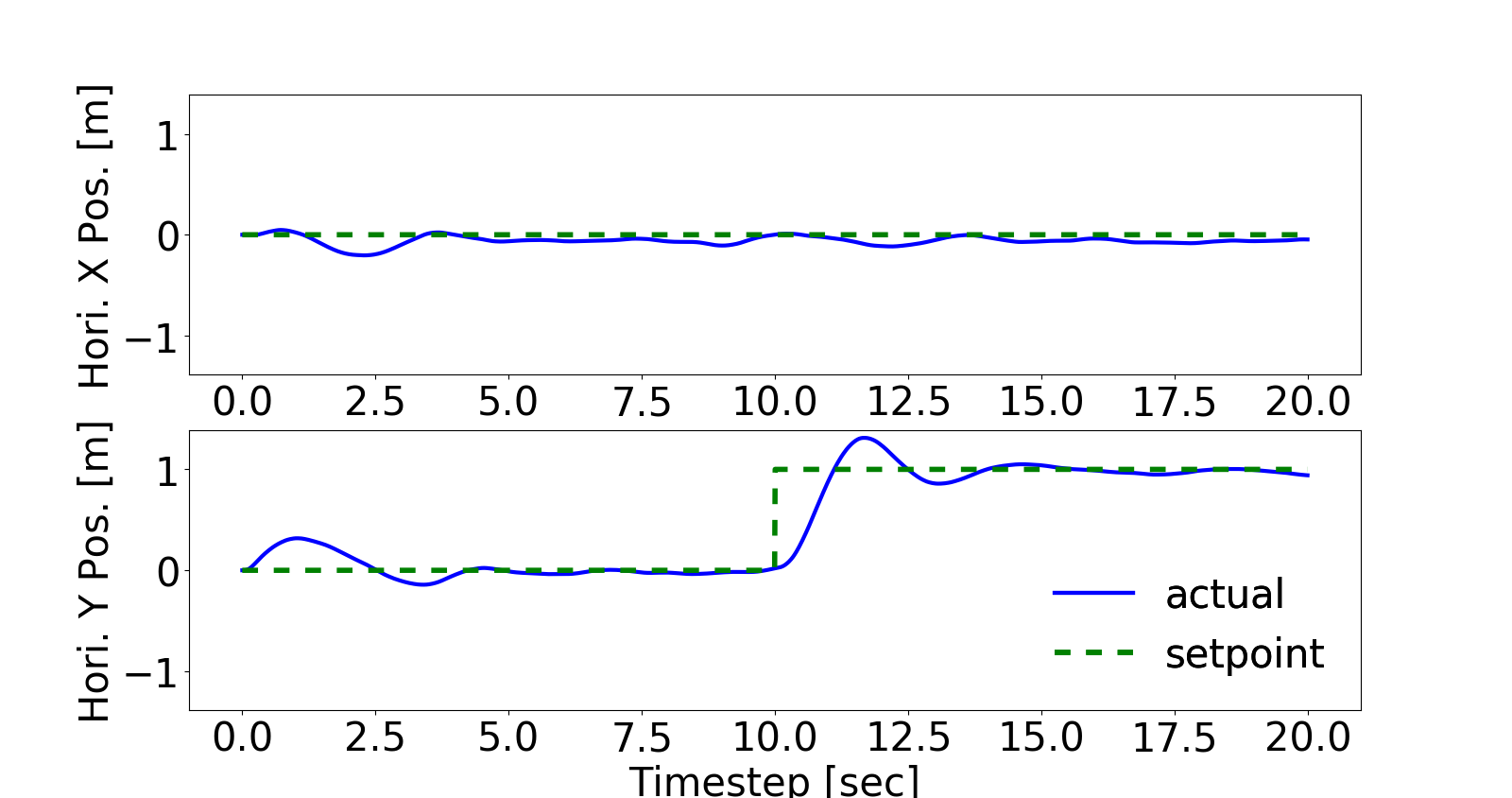}
  \label{fig:4_xy}
\end{subfigure}
\\[-20pt]
\begin{subfigure}{\textwidth}
  \centering
  \includegraphics[width=\linewidth]{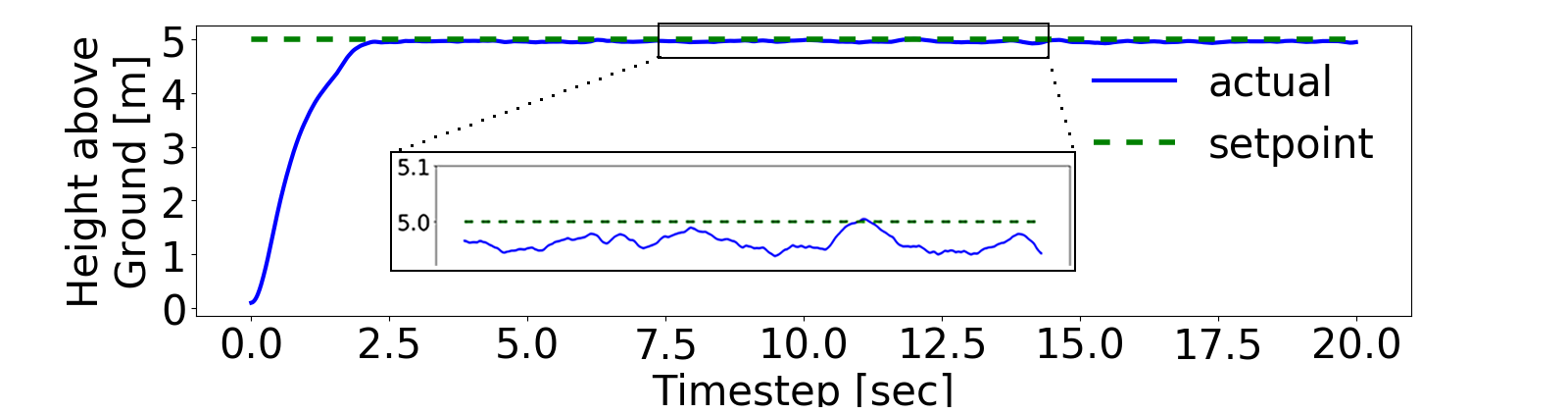}
  \label{fig:4_z}
\end{subfigure}
\\[-8pt]
\begin{subfigure}{\textwidth}
  \centering
  \includegraphics[width=\linewidth,trim={0cm 0cm 0cm 1.cm},clip]{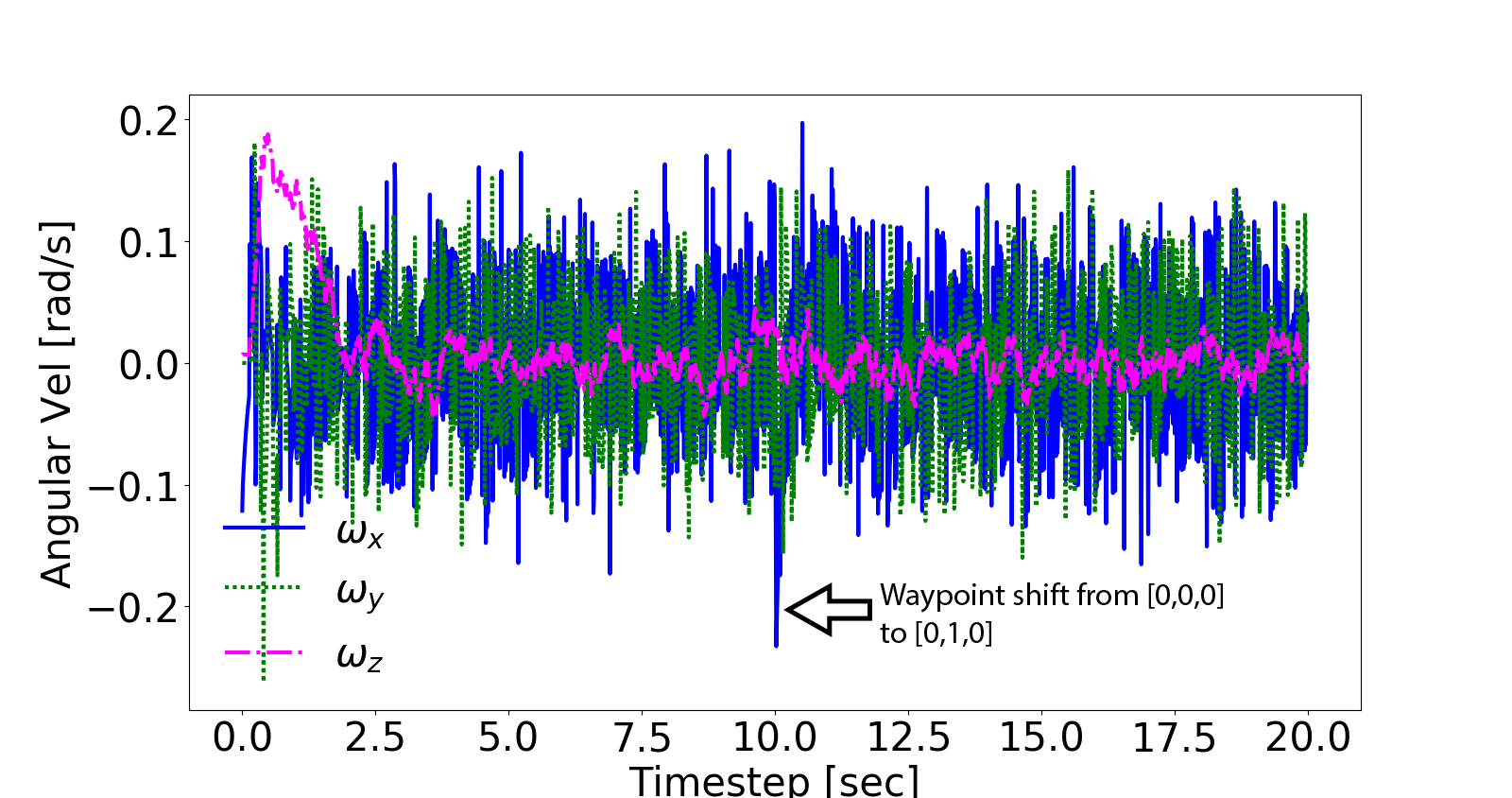}
  \label{fig:4_pqr}
\end{subfigure}
\vspace{-0.6\baselineskip}
\caption{Quadcopter with no propeller loss. Zoomed height plot is also shown.}
\label{fig:4_prop}
\end{minipage}
\end{figure}

\section{Results}\label{section:result}
We evaluated the performance of the 3 controllers individually, as well as in combination, while the quadcopter is performing waypoint tracking. We plotted its position $(x,y,z)$ and angular velocity $(w_{x}, w_{y}, w_{z})$ to demonstrate waypoint tracking and stability, respectively. The quadcopter is initialized at the origin and is directed to reach height $(z)$ of 5 m, simulating a take-off. After 10 seconds, the target-waypoint is shifted by 1 m in the positive Y direction. In the propeller loss scenarios, the propeller is turned off manually and time taken to regain stability is calculated.

\begin{figure}[t]
\begin{minipage}{.5\textwidth}
\begin{subfigure}{\textwidth}
  \centering
  \includegraphics[width=\linewidth]{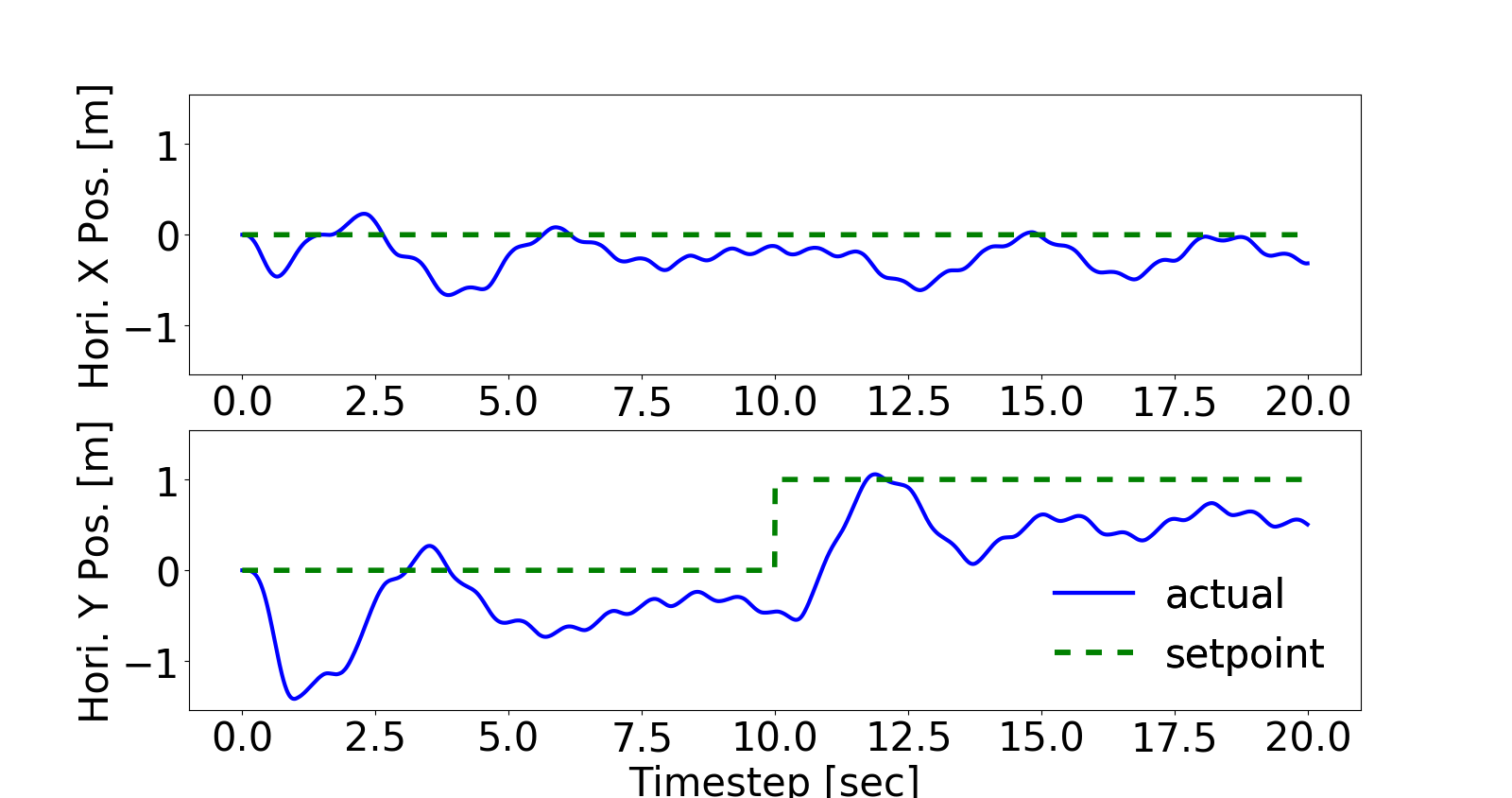}
  \label{fig:3_xy}
\end{subfigure}
\\[-20pt]
\begin{subfigure}{\textwidth}
  \centering
  \includegraphics[width=\linewidth]{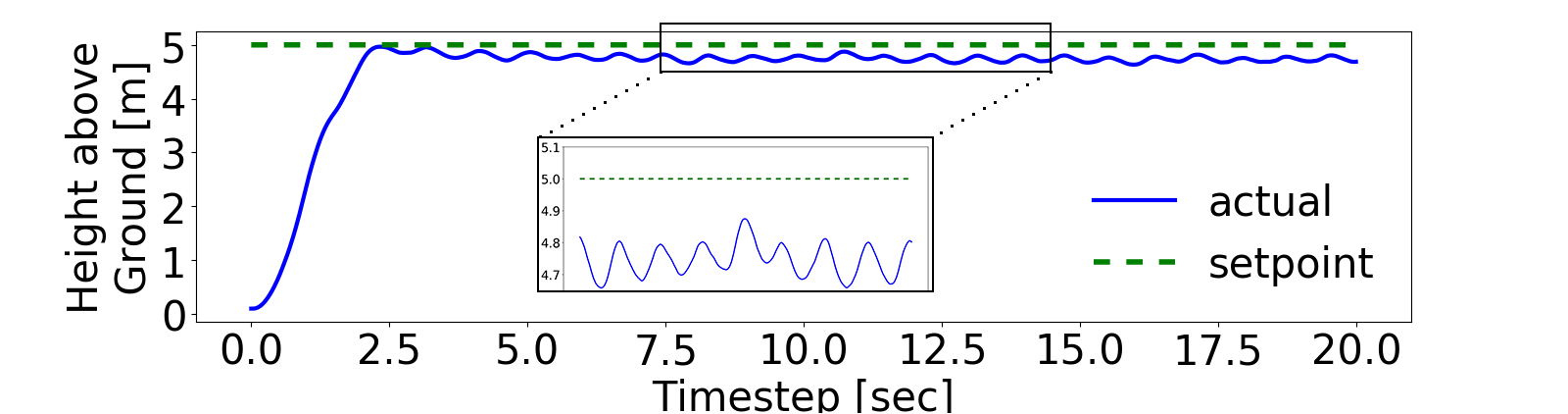}
  \label{fig:3_z}
\end{subfigure}
\\[-12pt]
\begin{subfigure}{\textwidth}
  \centering
  \includegraphics[width=\linewidth,trim={0cm 0cm 0cm 1.cm},clip]{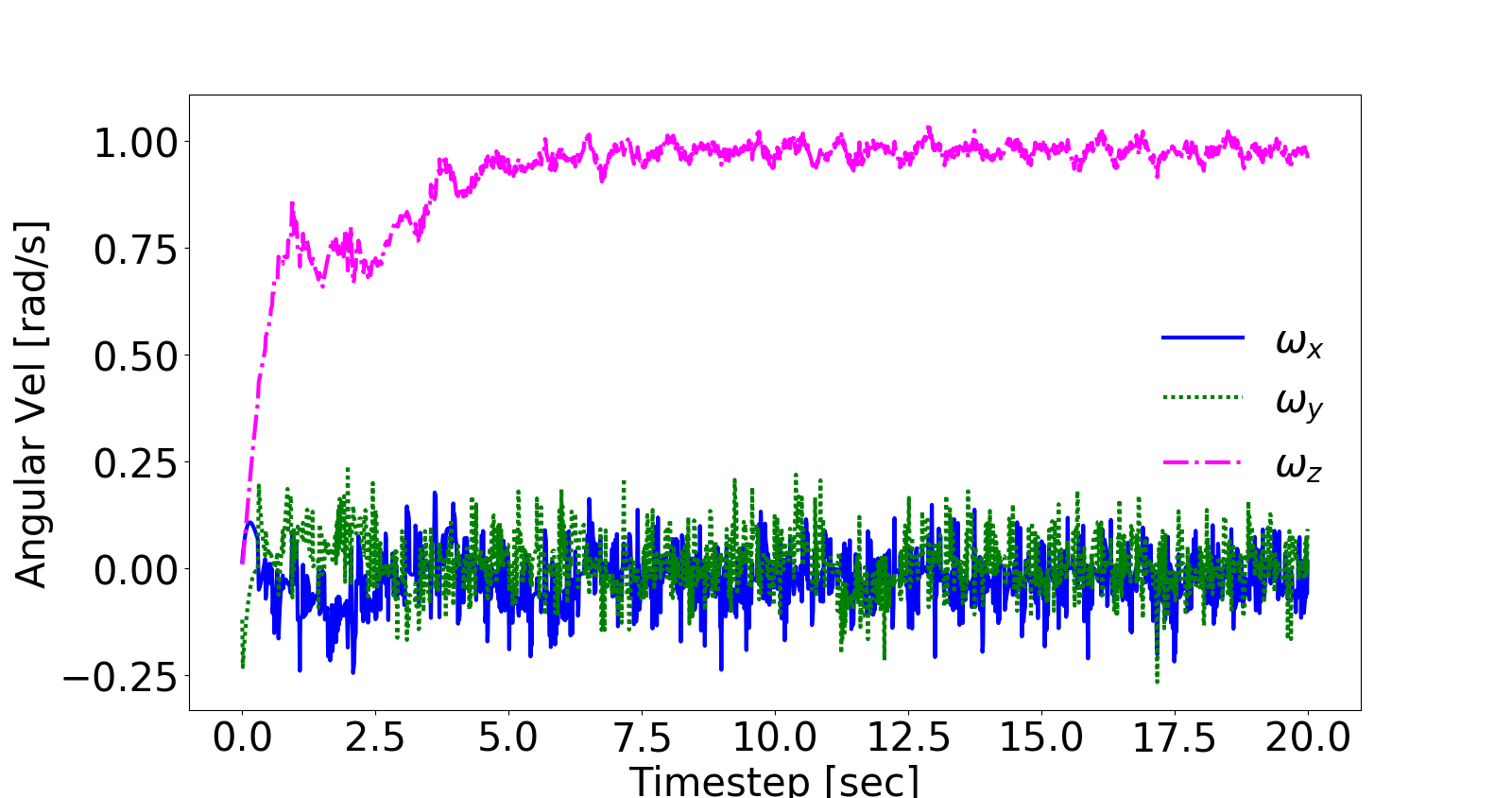}
  \label{fig:3_pqr}
\end{subfigure}
\vspace{-0.6\baselineskip}
\caption{One propeller lost.}
\label{fig:3_prop}
\end{minipage}
\end{figure}

\subsection{Implementation for no propeller loss case}
As can be seen from Fig. \ref{fig:4_prop}, the policy is able to track the waypoint accurately while keeping the quadcopter stable with near $0$ angular velocities. It also accommodates the waypoint shift. These results are comparable with \cite{RL_main}.

\begin{figure}
\begin{minipage}{.5\textwidth}
\begin{subfigure}{\textwidth}
  \centering
  \includegraphics[width=\linewidth]{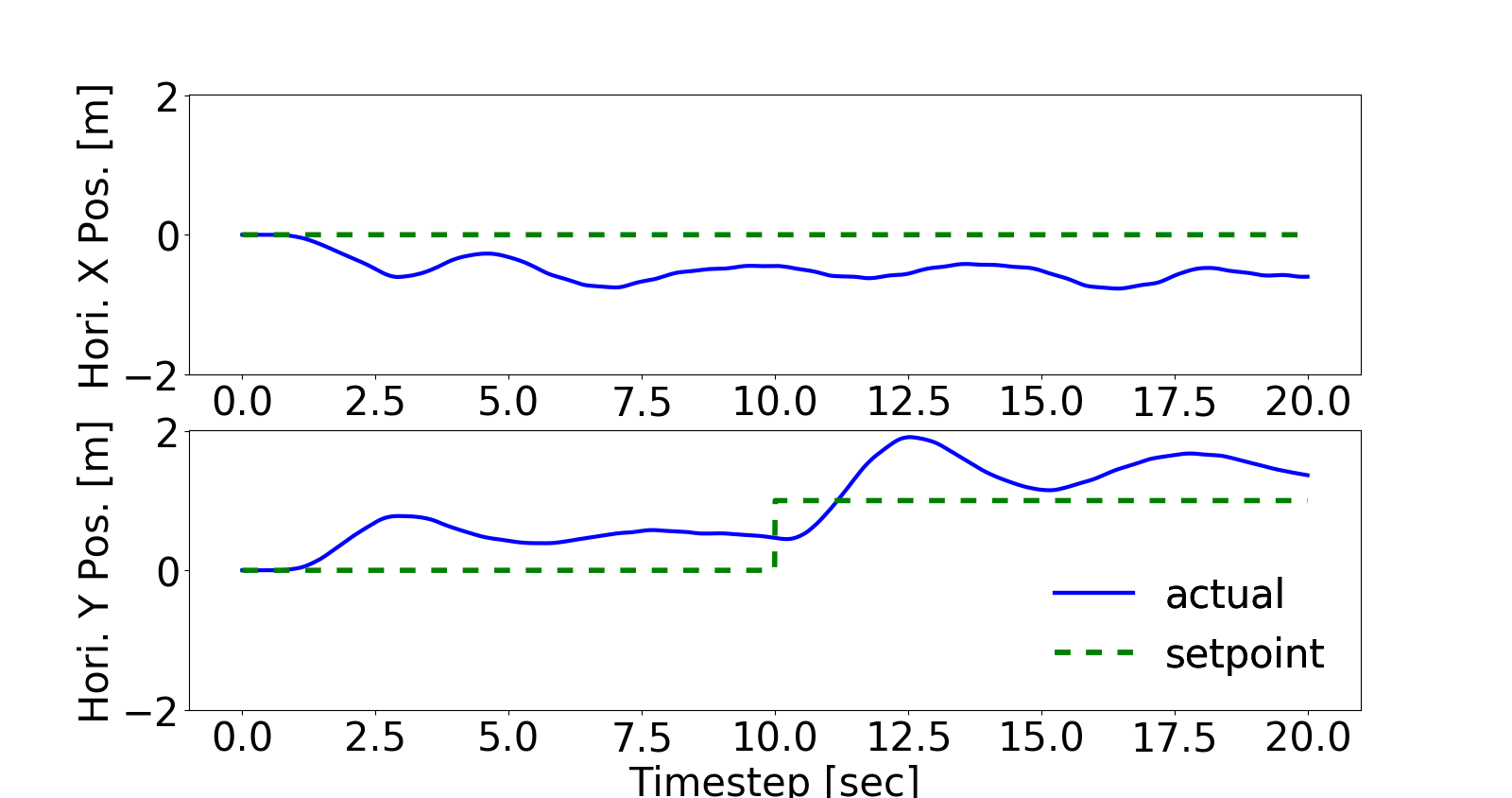}
  \label{fig:2_xy}
\end{subfigure}
\\[-20pt]
\begin{subfigure}{\textwidth}
  \centering
  \includegraphics[width=\linewidth]{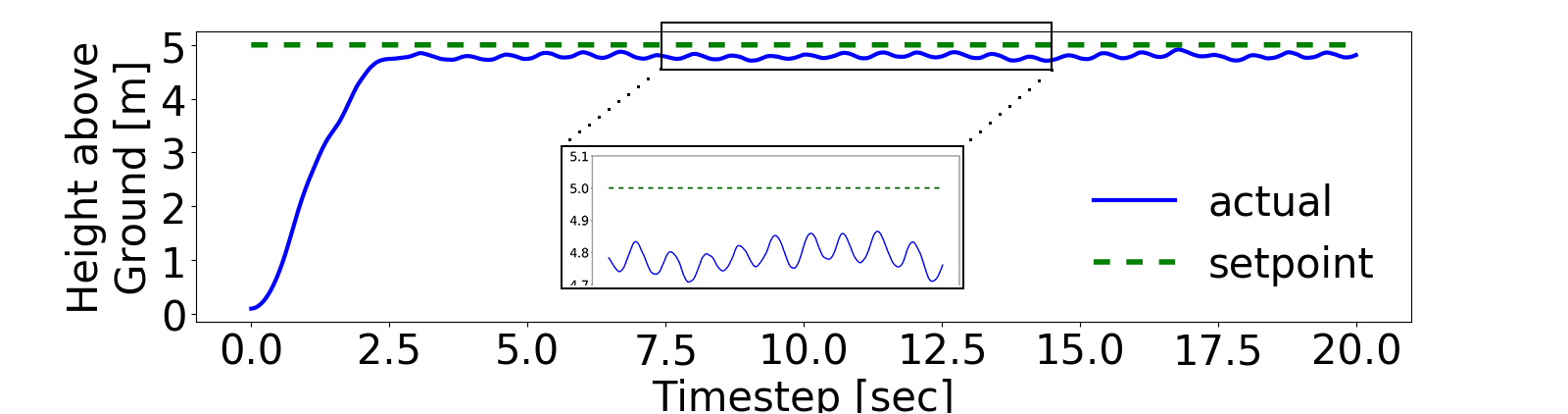}
  \label{fig:2_z}
\end{subfigure}
\\[-12pt]
\begin{subfigure}{\textwidth}
  \centering
  \includegraphics[width=\linewidth,trim={0cm 0cm 0cm 1.cm},clip]{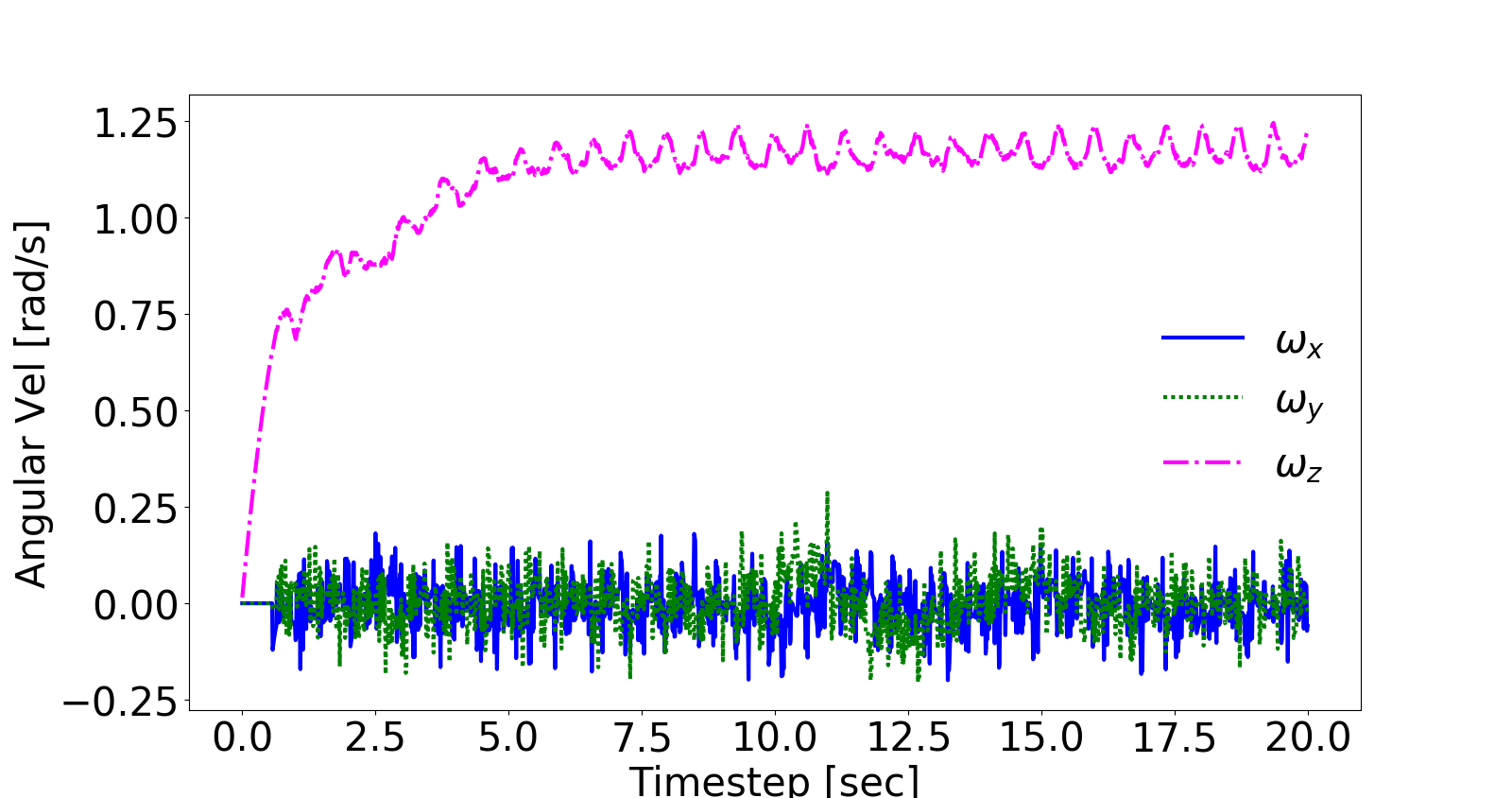}
  \label{fig:2_pqr}
\end{subfigure}
\vspace{-0.6\baselineskip}
\caption{Two propellers lost.}
\label{fig:2_prop}
\end{minipage}
\end{figure}

\reversemarginpar
\begin{figure}[t]
\begin{minipage}{.5\textwidth}
\begin{subfigure}{\textwidth}
  \centering
  \includegraphics[width=\linewidth]{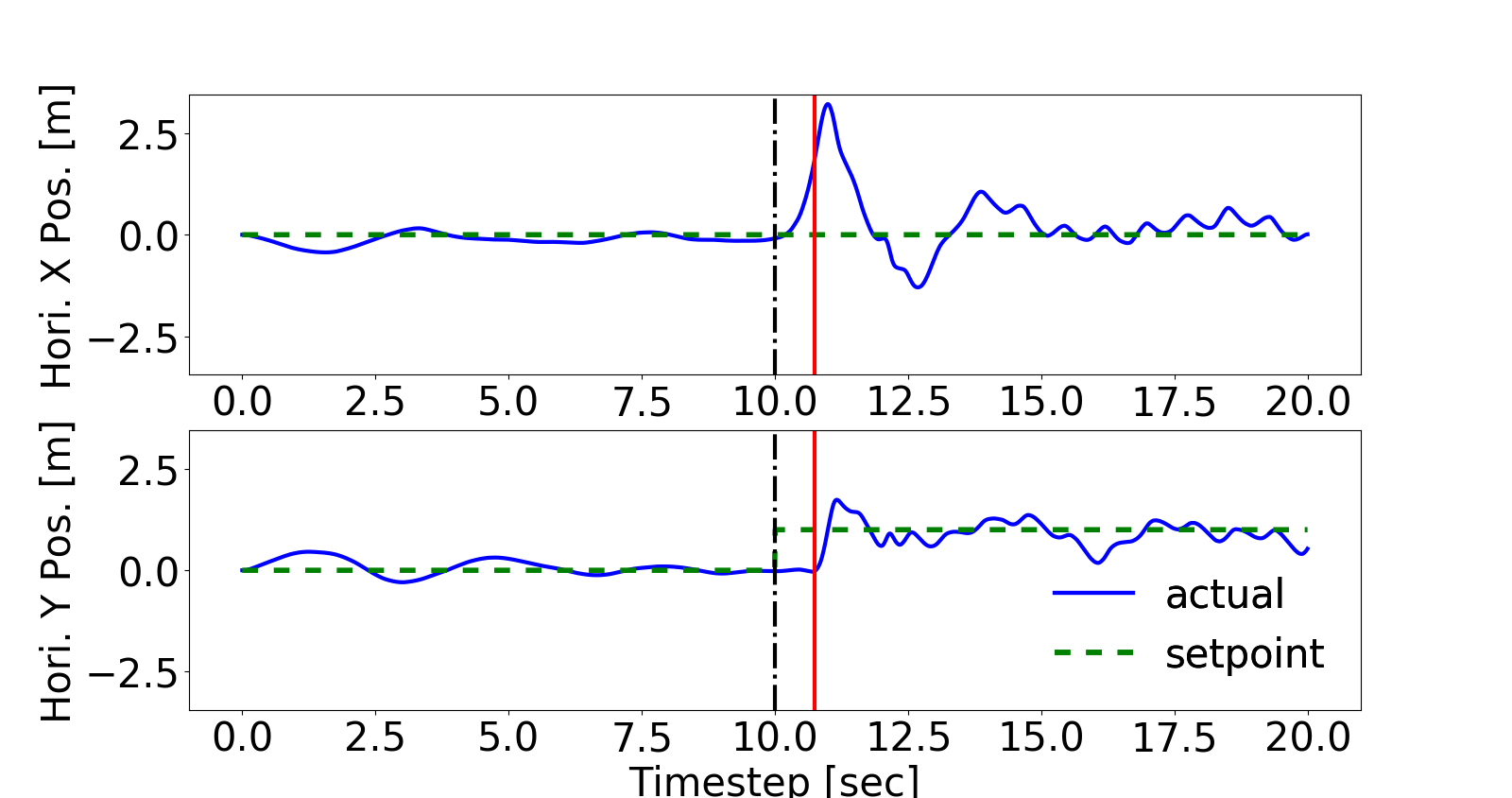}
  \label{fig:4_3_xy}
\end{subfigure}
\\[-20pt]
\begin{subfigure}{\textwidth}
  \centering
  \includegraphics[width=\linewidth]{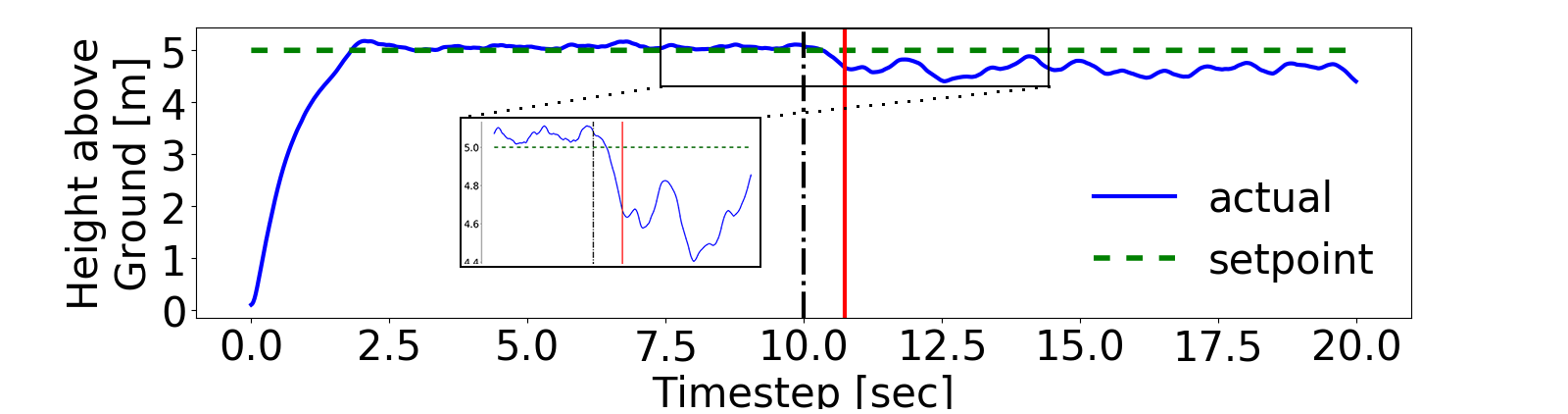}
  \label{fig:4_3_z}
\end{subfigure}
\\[-12pt]
\begin{subfigure}{\textwidth}
  \centering
  \includegraphics[width=\linewidth,trim={0cm 0cm 0cm 1.0cm},clip]{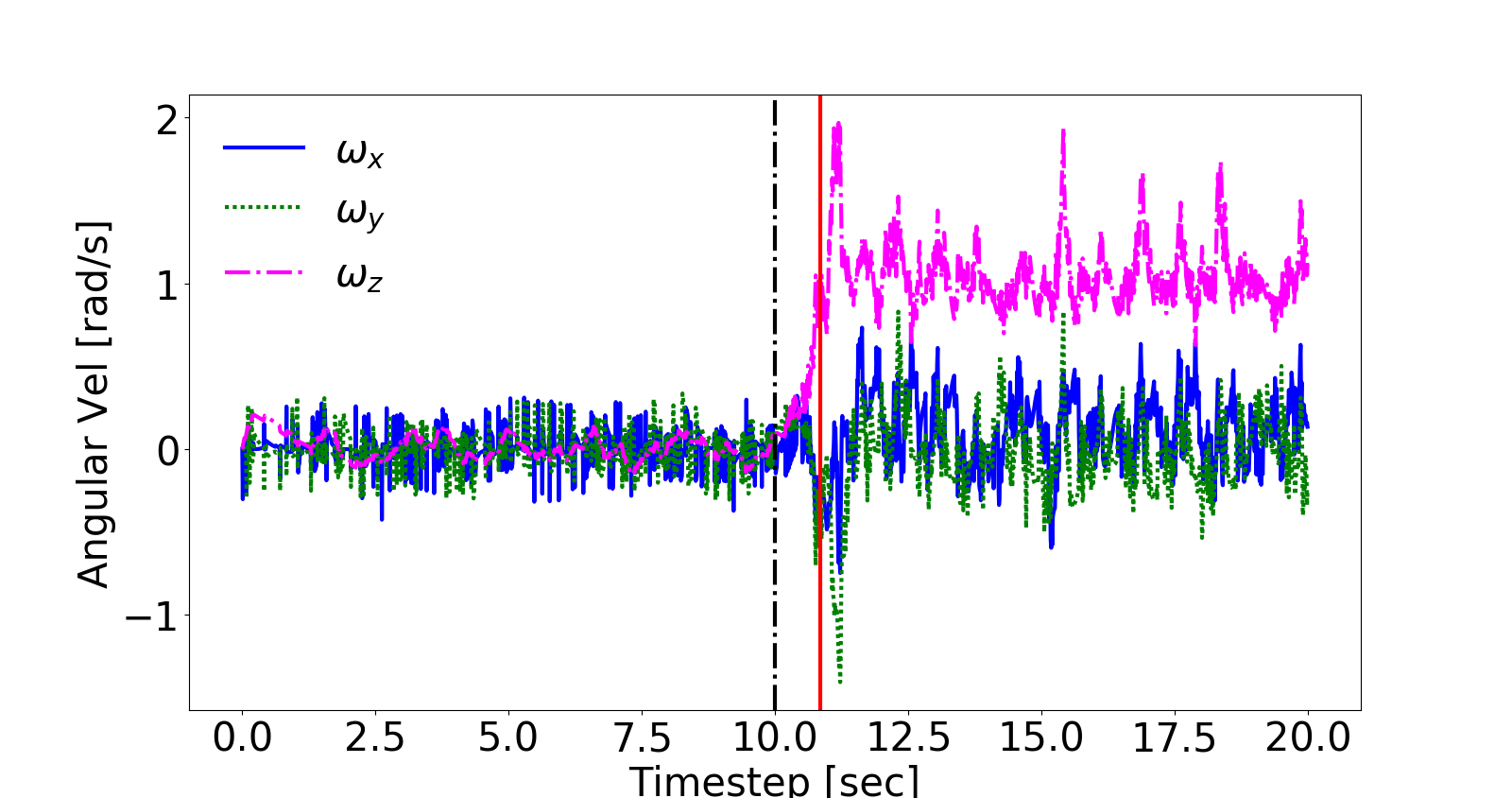}
  \label{fig:4_3_pqr}
\end{subfigure}
\vspace{-0.6\baselineskip}
\caption{First propeller lost in mid-flight.}
\label{fig:4_3_prop}
\end{minipage}
\end{figure}

\subsection{Implementation for one propeller loss case}
As shown in \cite{control_main}, on losing a single propeller, the quadcopter loses one degree of freedom and to maintain stability, a non-zero angular velocity about the vertical axis has to be enforced. Fig. \ref{fig:3_prop} shows the position and angular velocity of the quadcopter when it takes off with a single broken propeller. As can be seen from the graphs, the quadcopter has a constant yaw rate of approximately 1.0 rad/s. The graphs also demonstrate that even after losing a propeller, the quadcopter is able to track the waypoint shift occurring at 10 seconds into the simulation. There is a constant offset in position which is solved as described in subsection \ref{subsec:off}. 

{  Comparing with \cite{control_main}, we observe much less oscillations in the X position, and much less angular frequency along the Z-axis, in our case (see Fig. \ref{fig:3_prop}), as against those shown in Fig. 4 of \cite{control_main}. We also demonstrate taking off with one failed propeller. The maximum errors in waypoint tracking for \cite{control_main} and our quadcopter are tabulated in Table \ref{tab:one_lost_tab}.}

\begin{table}
  \centering
  \resizebox{\columnwidth}{!}{%
    \begin{tabular}{|c|c|c|c|c|}
        \hline
        & \multicolumn{2}{c|}{1-prop lost} & \multicolumn{2}{c|}{2-prop lost} \\
        \hline
        Parameter & Ours & Ref.  \cite{control_main} & Ours & Ref. \cite{control_main}\\
        \hline
        Horizontal Error (X) (m) & 0.7 & 0.2 & 1 & 0.2\\
        \hline
        Horizontal Error (Y) (m) & 1.5 & 0.5 & 1 & 0.5\\
        \hline 
        Height Error (Z) (m) & 0.4 & 1 & 0.4 & 2.4\\
        \hline
        Angular Velocity (Z) (rad/s) & 1 & 13.4 & 1.2 & $>$30\\
        \hline
    \end{tabular}
    }
     \caption{{  Comparison with \cite{control_main} for one and two propeller failed quadcopter.}}
    \label{tab:one_lost_tab}
\end{table}

\subsection{Implementation for two propeller loss case}
Fig. \ref{fig:2_prop} shows results for the loss of two propellers on a quadcopter system. There is a constant yaw rate of around 1.2 rad $\text{s}^{-1}$, which is quite close to the yaw rate of one propeller-lost system. The offset in position is larger compared to the single propeller-lost case, but  can be solved using the method described in subsection \ref{subsec:off}. The graphs demonstrate that the quadcopter is stable and is able to carry out waypoint tracking even with waypoint shift. We can, therefore, adjust the target waypoints in a similar manner to perform soft landing in real-life situations. 

{  Doing a similar comparison between Fig. 8 and \cite{control_main}, we again see high frequency oscillations, and high angular frequency along the Z-axis, in Fig. 5 of \cite{control_main}. We also demonstrate taking off with two failed propellers. The maximum errors in waypoint tracking for \cite{control_main} and our quadcopter are tabulated in Table \ref{tab:one_lost_tab}.}

\subsection{Comparison of the quadcopter dynamics}

{ We would like to point out that the vehicle data in \cite{control_main}, is similar to a large extent with our quadcopter making the comparison feasible, although they do differ in some aspects. For the sake of completion, this data is given below with data from \cite{control_main} given in parentheses. Mass: 0.4 (0.5) kg; Arm length: 0.17 (0.17) m; Drag coefficient: 16x10$^{-3}$ (2.75x10$^{-3}$) N m s rad$^{-1}$; I$_{XX}$ and I$_{YY}$: 7x10$^{-3}$ (3.2x10$^{-3}$) Kg m$^{-2}$; I$_{ZZ}$: 12x10$^{-3}$ (5.5x10$^{-3}$) Kg m$^{-2}$.}

%
    
\normalmarginpar
\begin{figure}[t]
\begin{minipage}{.5\textwidth}
\begin{subfigure}{\textwidth}
  \centering
  \includegraphics[width=\linewidth]{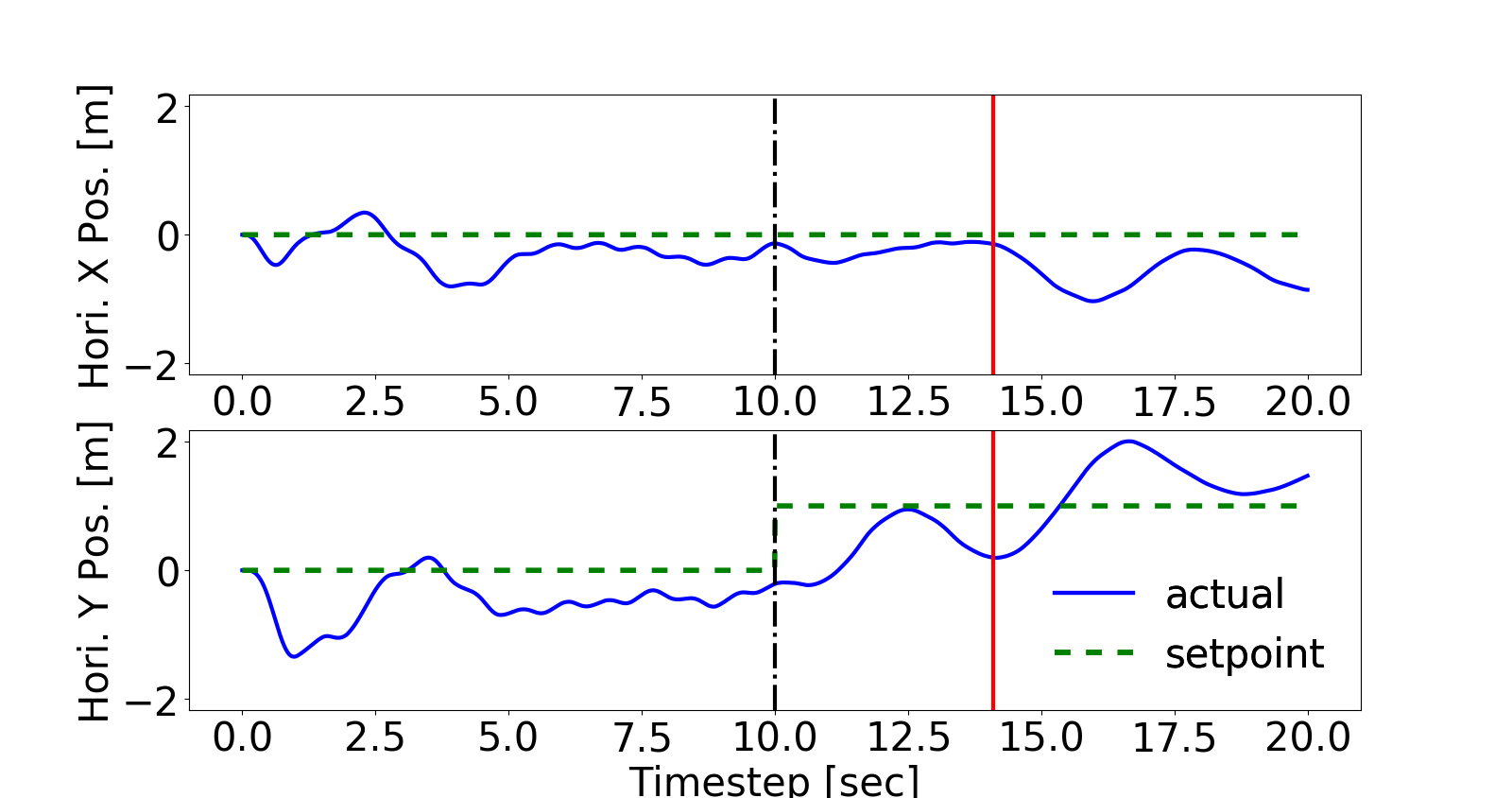}
  \label{fig:3_2_xy}
\end{subfigure}
\\[-20pt]
\begin{subfigure}{\textwidth}
  \centering
  \includegraphics[width=\linewidth]{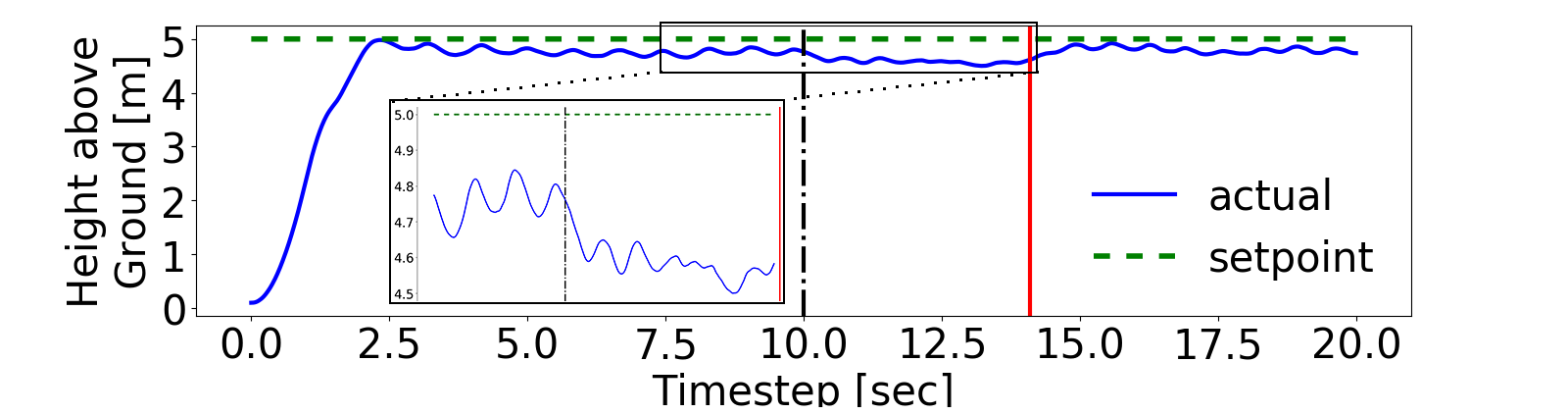}
  \label{fig:3_2_z}
\end{subfigure}
\\[-12pt]
\begin{subfigure}{\textwidth}
  \centering
  \includegraphics[width=\linewidth,trim={0cm 0cm 0cm 1.cm},clip]{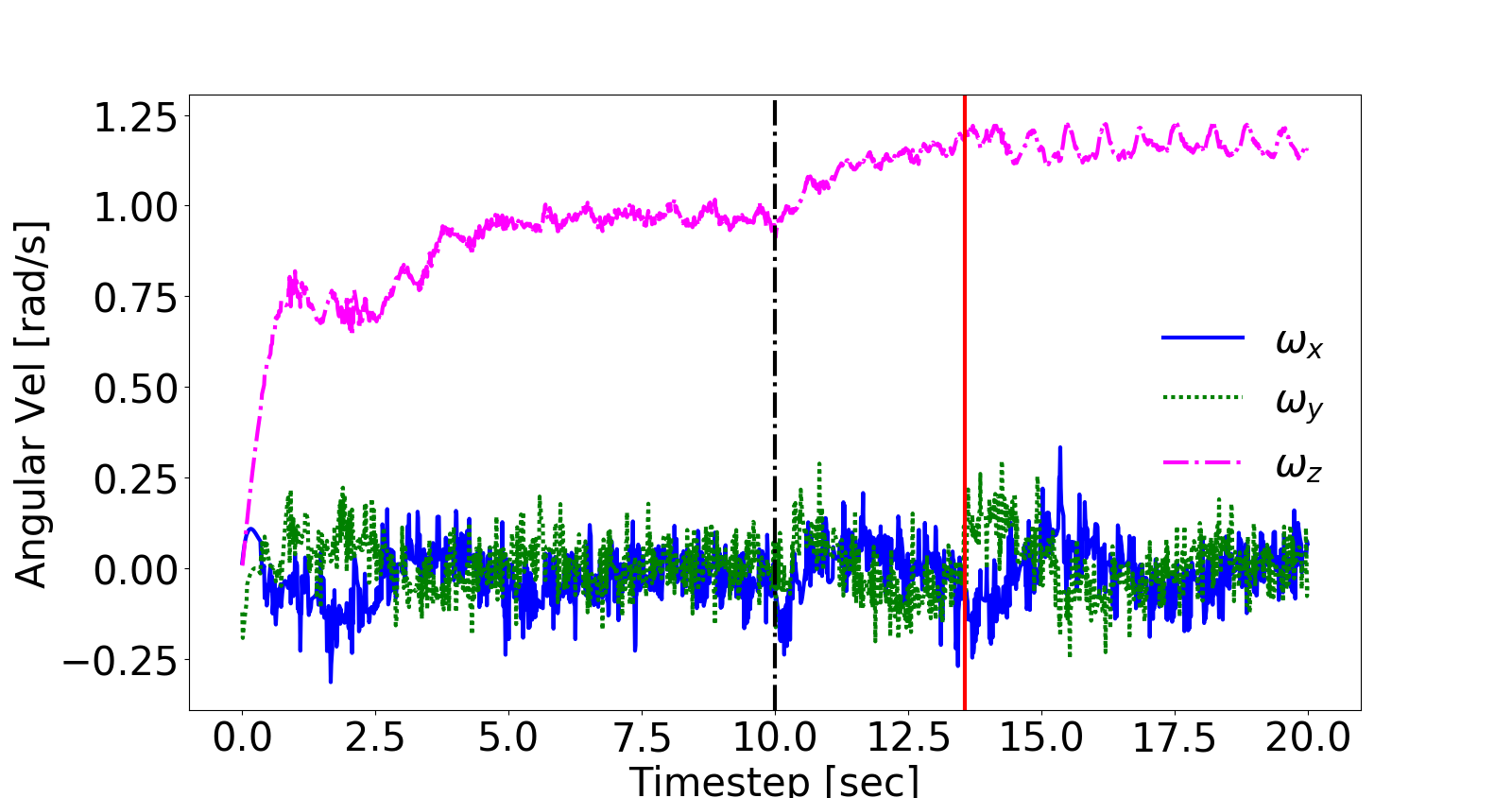}
  \label{fig:3_2_pqr}
\end{subfigure}
\vspace{-0.6\baselineskip}
\caption{Three propeller quadcopter with second propel-\\ler (opposing) lost in mid-flight.}
\label{fig:3_2_prop}
\end{minipage}
\end{figure}

\subsection{Integrating fault detection with the control agents}
\subsubsection{Transition from 4 working propellers to 3 working propellers}
Fig. \ref{fig:4_3_prop} shows the behavior of the quadcopter when a propeller fails mid-flight. The FD system identifies the failed propeller and switches to the appropriate control agent. The 2 vertical lines in the graphs represents the actual time at which the propeller failed and time at which the fault was detected, respectively. The graph shows a 1 second delay between the failure and its detection. Table \ref{tab:detect_time} shows the average time of detection for 5 runs with failure occurring at random timesteps.

\begin{table}
  \centering
    \begin{tabular}{|c|c|c|}
        \hline
        Propellers Lost & 1st & 2nd\\
        \hline
        Time (secs) & 0.75 & 4.07\\
        \hline
    \end{tabular}
    \caption{Average time taken to detect propeller failures}
        \label{tab:detect_time}
\end{table}

\subsubsection{Transition from 3 working propellers to 2 working propellers}
Fig. \ref{fig:3_2_prop} shows the behavior of the quadcopter when the second propeller also fails mid-flight. The $3\rightarrow2$ FD system kicks in and identifies the failed propeller and switches to the appropriate control agent. The two vertical lines in the graphs represents the actual time at which the propeller failed and the time at which the fault was detected, respectively. As can be seen, there is a delay of around 2.24 seconds between the propeller failure and its detection. From Table \ref{tab:detect_time}, we can also see the average time of detection for 5 runs, in which the failure occurs at random timesteps.

\begin{table}
  \centering
  \resizebox{\columnwidth}{!}{%
    \begin{tabular}{|p{0.6\columnwidth}|p{0.3\columnwidth}|}
        \hline
        & Failure Rate (\%) \\
        \hline
        Control agent - No propeller loss & 2 \\
        \hline
        Control agent - One propeller loss & 24.2 \\
        \hline
        Control agent - Two propeller loss & 36 \\
        \hline
    \end{tabular}
    }
    \caption{The failure rate (in 500 runs) for independent control agents and various propeller loss cases.}
        \label{tab:fail_rate}
\end{table}

\begin{figure}
  \centering
  \includegraphics[width=0.75\linewidth,]{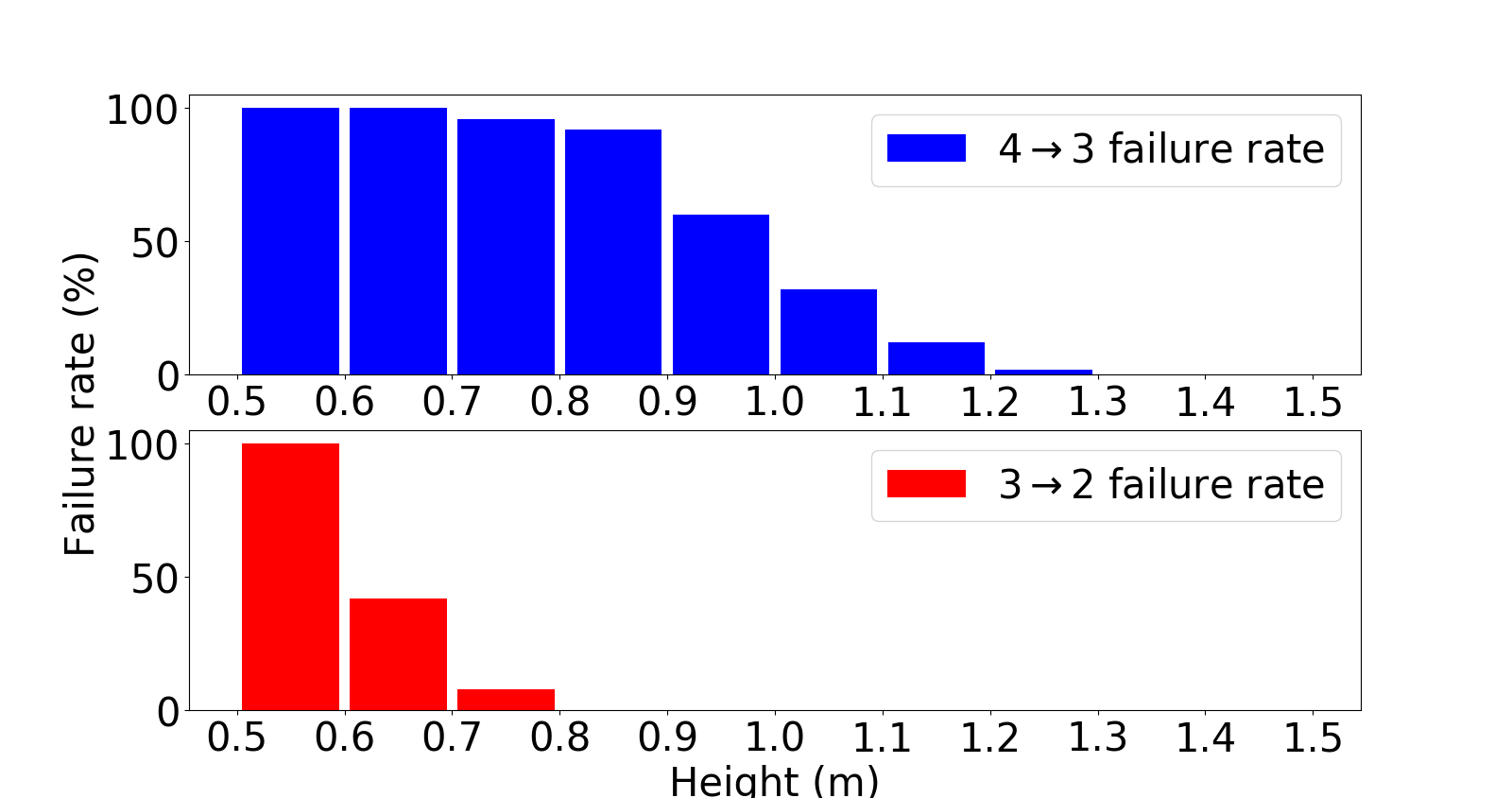}
  \label{fig:fail_43_lol}
\caption{Failure rate of $4\rightarrow3$ and $3\rightarrow2$ in 500 runs spread across a height range of 1 meter.}
\label{fig:fail_both}
\end{figure}

\subsection{Failure rate calculation}
In this paper, the failure condition for the quadcopter is whenever it hits the ground. The failure rate is recorded from 500 runs with random initialization of orientation, position, linear and angular velocity. The orientation was sampled uniformly in $SO(3)$ and the other quantities were sampled uniformly in $[-1,1]$. For the 4, 3, and 2 propeller scenarios, the target height was fixed at 5 meters above the ground. The results are given in Table \ref{tab:fail_rate}. The random initialization in the isolated 4, 3, and 2 propeller scenarios caused some quadcopters to start in an irrecoverable state, for example, upside down or very high linear velocity towards the ground and so on. Recovery from these states becomes harder due to the lost degree-of-freedom and insufficient thrust, thus increasing the failure rate. 

Fig. \ref{fig:fail_both} shows the failure rate when propellers are lost mid flight from both $4\rightarrow3$ and $3\rightarrow2$ cases. The target height of the quadcopter in the 500 runs was evenly distributed between 0.5$m$-1.5$m$ for both $4\rightarrow3$ and $3\rightarrow2$ scenarios. These heights were chosen based on Figs. \ref{fig:4_3_prop} and \ref{fig:3_2_prop}, which show an approximate drop in height of 0.5m for both first and second propeller failure. From Fig. \ref{fig:fail_both}, we are able to find a height threshold at which the failure recovery system is not able to switch in time and stop the quadcopter from crashing to the ground. \href{https://youtu.be/3KF4GtAux00}{Videos}\footnote{ \normalmarginpar \href{https://youtu.be/3KF4GtAux00}{https://youtu.be/3KF4GtAux00}} showing the experiments in detail and the implementation\href{https://github.com/Aakriti05/Prop-Fail-Detect-Control-RL}{code}\footnote{\href{https://github.com/Aakriti05/Prop-Fail-Detect-Control-RL}{https://github.com/Aakriti05/Prop-Fail-Detect-Control-RL}} are available in the given link.

\section{CONCLUSIONS}\label{section:conc}
In this paper, we have proposed a system for mid-flight failure detection and control in case of multiple propeller loss in a quadcopter. Firstly, we showed how RL agents can learn to control quadcopters with 0, 1 and 2 (opposing) propeller(s) lost. We showed that the quadcopter learned to do waypoint tracking while maintaining stability, even with 1 and 2 (opposing) propeller(s) failed. Secondly, we developed a novel FD system using deep learning which can detect the propeller(s) failure and switch to the appropriate controller. This method requires only the previous states of the quadcopter and is able to detect the propeller loss within 2.5 seconds, thus removing the need and maintenance for any additional sensor hardware on the quadcopter. We have also shown, in simulation, that the detection and switching can happen in real-time, preventing the quadcopter from crashing and enabling it to either land or continue its mission.

Future scope of this work can be to replace the inner loop based on PD controller, with an RL agent to make the whole system model-free, and completely discard the need to develop a mathematical model of the quadcopter. {  One can also use transfer-learning for training the controllers for propeller loss. This may allow training the other controllers using less number of trajectories.} The implementation of this system on a physical quadcopter is the next step of this research work. {  This can be done by transferring the weights from simulation to physical quadcopter where it would be trained. A motion capture system can ensure fairly  accurate values of position and orientation for training the network. }


\addtolength{\textheight}{-10cm}   


\bibliographystyle{IEEEtran}
\bibliography{IEEEexample}
\end{document}